\documentclass[letterpaper, 10 pt, conference]{ieeeconf}  

\IEEEoverridecommandlockouts                              

\usepackage{graphics}
\usepackage{epsfig} 
\usepackage{mathptmx} 
\usepackage{times} 
\usepackage{amsmath} 
\usepackage{amssymb}  
\usepackage{multirow}
\usepackage{url}
\usepackage{caption}
\usepackage{subcaption}
\usepackage[table]{xcolor}
\usepackage{import}
\usepackage{hyperref}

\usepackage{booktabs}
\usepackage{tabularx}
\usepackage{microtype}
\usepackage{gensymb} 
\usepackage{cite}
\usepackage{stfloats}

\usepackage{bm}
\usepackage{siunitx}
\title{\LARGE \bf
 Safety-Conscious Pushing on Diverse Oriented Surfaces with Underactuated Aerial Vehicles}

\author{ Tong Hui$^{*}$, Manuel J. Fernández González$^{*}$, Matteo Fumagalli
\thanks{This work has been supported by the European Unions Horizon 2020 Research and Innovation Programme AERO-TRAIN under Grant Agreement No. 953454. $^{*}$ The authors equally contribute to the work.}
\thanks{All authors are with Department of Electrical and Photonics Engineering, Technical University of Denmark. Corresponding author email:
        {\tt\small tonhu@dtu.dk}}
}

\begin{document}

\maketitle
\thispagestyle{empty}
\pagestyle{empty}

\begin{abstract}
Pushing tasks performed by aerial manipulators can be used for contact-based industrial inspections. Underactuated aerial vehicles are widely employed in aerial manipulation due to their widespread availability and relatively low cost. Industrial infrastructures often consist of diverse oriented work surfaces.
When interacting with such surfaces, the coupled gravity compensation and interaction force generation of underactuated aerial vehicles can present the potential challenge of near-saturation operations. The blind utilization of these platforms for such tasks can lead to instability and accidents, creating unsafe operating conditions and potentially damaging the platform.
In order to ensure safe pushing on these surfaces while managing platform saturation, this work establishes a safety assessment process. This process involves the prediction of the saturation level of each actuator during pushing across variable surface orientations. Furthermore, the assessment results are used to plan and execute physical experiments, ensuring safe operations and preventing platform damage.
\end{abstract}

\section{INTRODUCTION}\label{intro}
Recent studies have shown significant growth in the successful integration of aerial manipulation into industrial applications ~\cite{Anibal2021}. Pushing tasks performed by aerial manipulators at height can be used for contact-based inspections to prevent hazardous working conditions for human workers. This has motivated studies on the use of aerial vehicles for such operations. Among the developed aerial manipulation platforms for pushing tasks, underactuated aerial vehicles have gained widespread popularity as flight platforms due to constructional reasons~\cite{Anibal2021,emran}, e.g. high availability in the market and relatively low cost.

Aerial robotic systems are classified as floating-based systems. In conventional fixed-based systems, it is often assumed that the connected ground can effectively generate infinite reaction forces and torques (i.e. wrenches) when the environment interacts with the robotic system~\cite{Siciliano,float}. However, in the context of floating-based systems, this assumption no longer holds. In these systems, the actuation of such systems becomes pivotal, responsible for supplying the necessary wrenches during interactions with the environment. For an underactuated floating-based system, two key factors come into play: (1) the saturation of the actuators, (2) the coupling between linear and angular dynamics~\cite{rashad2020,dario2018}. These factors collectively limit the magnitude and direction of the exerted wrenches that the robotic system can apply to the environment during physical interactions~\cite{float}. 

Industrial infrastructures are often composed of diverse oriented work surfaces. To achieve pushing tasks with underactuated aerial vehicles on such surfaces, manipulators with at least 1-DoF (Degree of Freedom) are often attached to the aerial vehicle~\cite{byun2023, hui2023}. Research efforts have been made on interaction control for such operations with underactuated aerial vehicles~\cite{fumagalli2012, lipp2018, lipp2012, tognon2016, hui2023, sch2013, byun2023}. When interacting with diverse oriented surfaces, the coupled gravity compensation and interaction force generation of underactuated aerial vehicles present the potential challenge of near-saturation operations. The blind utilization of these platforms for such tasks can result in instability and accidents, leading to unsafe operating conditions and platform damage. These issues underscore the importance of identifying system limitations in such operations.

In the work by Lassen et al.~\cite{lassen}, they were among the first to identify an operational envelope, encompassing pitch angle and thrust force parameters, necessary for an underactuated aerial vehicle to maintain stable pushing with a flat vertical surface. In our previous work~\cite{hui2023}, a static-equilibrium based force modeling approach is presented regarding to pushing on diverse oriented work surfaces. The singularity is outlined via predicting the total platform thrust using the derived force models when interacting with surfaces at different orientation. The singularity highlights the limitations in exerting feasible interaction force magnitude associated with variable surface orientation considering total system saturation. To ensure safe operations, it is however essential to 
guarantee that each individual actuator works normally within its saturation.
\begin{figure}[!t]
      \centering
\includegraphics[width=\columnwidth]{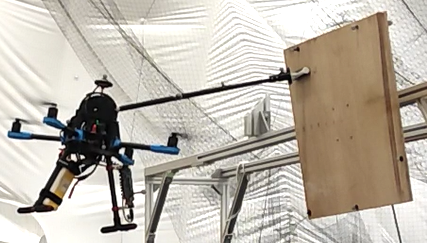}
      \caption{A quadrotor-based aerial manipulator is pushing on a flat oriented work surface.}
      \label{fig:action}
\end{figure}
\subsection{Main Contribution}\label{sec:contribution}
In order to ensure safe pushing on diverse oriented work surfaces with an underactuated aerial vehicle, this work establishes a safety assessment process based on the saturation level of each individual actuator during interactions. This process includes:
\begin{itemize}
    \item (I) identifying safe and critical operation zones during pushing;
    \item (II) assessing the risk level within the critical zone.
\end{itemize}
The safe interaction zone is defined as the region where all actuators operate under nominal conditions, while the critical zone involves at least one actuator operating near saturation. We apply this safety assessment process to a quadrotor-based aerial manipulator system with an 1-DoF link attached to it for a pushing task (see Fig.~\ref{fig:action}). This process is accomplished by comparing the predicted thrust of each actuator using force models derived in ~\cite{hui2023} with its saturation across different surface orientation. Furthermore, we use the safety assessment results to plan and conduct practical experiments, safely validating the force model proposed in~\cite{hui2023} and avoiding platform damage.

\section{Background}
\label{sec:bg}
In this section, we briefly introduce the force modeling approach for a pushing task on diverse oriented surfaces with an underactuated aerial vehicle presented in \cite{hui2023}. An aerial manipulator composed of a quadrotor and an 1-DoF link rigidly attached to it is used to push on flat oriented work surfaces. The rigid link acts as the manipulator with a single contact point at the EE (End-Effector) tip. The axis along the length of the EE link is called the interaction axis. For maintaining a stable pushing without slipping at the EE tip, it is desired to only exert a force along the normal vector of the work surface at the contact point, i.e. zero lateral friction forces.
\begin{figure}[t]
      \centering    \includegraphics[width=0.9\columnwidth]{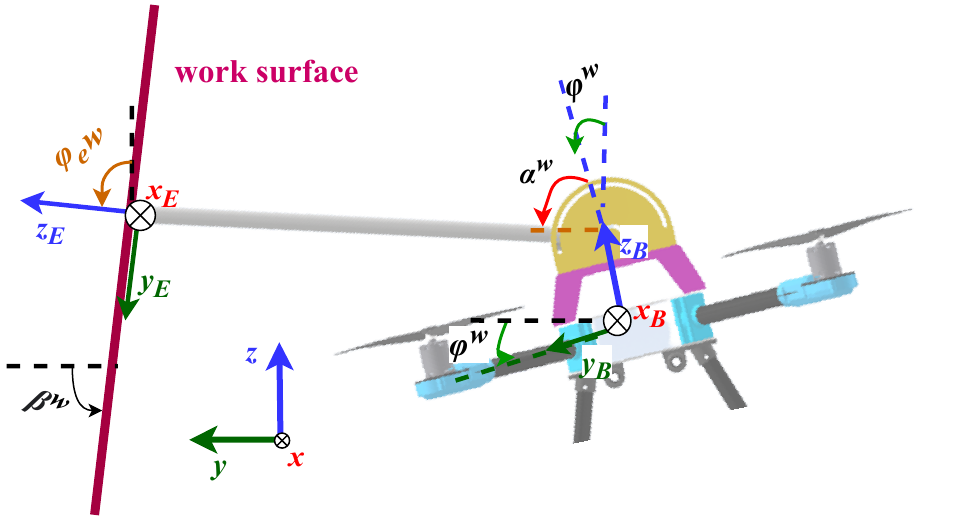}
      \caption{System model in 2-D plane $(\bm{y},\bm{z})$, coordinate frames, roll angle of the aerial vehicle $\varphi^{w}$, joint position $\alpha^{w}$, work surface orientation $\beta^{w}$.}
      \label{fig:system}
\end{figure}
To do so, the system is oriented such that the interaction axis coincides with the normal vector of the work surface. Therefore, without effecting the main contribution, we consider a simplified planar system in a 2-D (dimensional) plane that contains the normal vector of the work surface, as in Fig.~\ref{fig:system}.
\subsection{Notation} \label{sec:notation}
We denote $\mathcal{F}^w=\{O;\bm{x},\bm{y},\bm{z}\}$ as the inertial frame. $\mathcal{F}^B=\{O_B;\bm{x}_B,\bm{y}_B,\bm{z}_B\}$ represents the body frame attached to the CoM (Center of Mass) of the aerial vehicle, and $\mathcal{F}^E=\{O_E;\bm{x}_E,\bm{y}_E,\bm{z}_E\}$ is the EE frame attached to the contact point between the EE tip and the work surface.
The work surface orientation is resulted from rotating the plane $(\bm{x},\bm{y})$ of the inertial frame around $\bm{x}$ axis with an angle $\beta^{w}$, where $\beta^{w} \in [-\frac{\pi}{2},\frac{\pi}{2}]$ being positive when the rotation is anticlockwise. The rigid link is connected to the aerial vehicle by a revolute joint. $\alpha^{w}$ denotes the joint position expressed in the inertial frame while $\alpha^{w}=0$ when $\bm{z}_E$ aligns with $\bm{z}_B$. We define the relative orientation of the body frame w.r.t. (with respect to) the inertial frame around the axis $\bm{x}$ in plane $(\bm{y},\bm{z})$ as the roll angle $\varphi^{w}$ of the aerial vehicle expressed in the inertial frame. $\alpha^{w}, \varphi^{w} \in [-\frac{\pi}{2},\frac{\pi}{2}]$ are positive while rotating anticlockwise around $\bm{x}$ axis. In the restricted 2-D plane, the relative orientation of the EE frame $\mathcal{F}^E$ w.r.t. the inertial frame expressed in the inertial frame is given by: 
\begin{equation}
\varphi_e^{w}=\varphi^{w}+\alpha^{w}.
\label{eq:eeori}
\end{equation}
$\bm{F}_C^{w}=\begin{bmatrix}
 \bm{f}_C^{w}\\ \bm{\tau}_C^{w}
\end{bmatrix} \in \mathbb{R}^6$ represents the interaction wrench acting on the aerial manipulator from the environment expressed in the inertial frame. $\begin{bmatrix}
 \sum_{i=1}^4\mathbf{T}_i^{w}\\ \sum_{i=1}^4 \Gamma_i^{w}
\end{bmatrix} \in \mathbb{R}^6$ is the stack of thrust vector and torque vector generated by 4 propellers (or rotors) of the aerial vehicle.
Considering the simplified planar system model, one has:
\begin{align}
    &\bm{f}_C^{w}=\begin{bmatrix}
 0\\f_{C}^{w_y}\\f_{C}^{w_z} \end{bmatrix}, \bm{\tau}_C^{w}= \begin{bmatrix}
\tau_{C}^{w_x}\\0\\0
\end{bmatrix},\\
&\sum_{i=1}^4\mathbf{T}_i^{w}=\begin{bmatrix}
 0\\T_{a}^{w_y}\\T_{a}^{w_z} \end{bmatrix}, \sum_{i=1}^4 \Gamma_i^{w}= \begin{bmatrix}
\tau_{a}^{w_x}\\0\\0
\end{bmatrix}.
\end{align} 
\subsection{Task Constraints} \label{sec:ps}
The introduced aerial manipulator is subjected to execute the following targeted task: apply a force vector along positive $\bm{z}_E$ axis of the EE frame, directed towards a work surface that is perpendicular to it, while preserving stability. Assuming that the force vector is strictly perpendicular to the work surface and the friction force is negligible, the task introduces the following constraints on the system ~\cite{hui2023}:
\begin{subequations}\label{eq:task}
\begin{equation}  \beta^{w}=\varphi_e^{w}=\varphi^{w}+\alpha^{w},\label{eq:angles}
\end{equation}
\begin{equation}
    \bm{\upsilon}_e^{w}=\bm{0}_6,
\end{equation}
\begin{equation}
\bm{F}_C^{E}=\begin{bmatrix}
        0&0&f_e & \bm{0}_{3}
         \end{bmatrix}^{\top}, \ f_e \geq 0,
\end{equation}
\end{subequations}
where $\bm{F}_C^{E} \in \mathbb{R}^6$ represents the interaction wrench expressed in the EE frame $\mathcal{F}^E$, $\bm{\upsilon}_e^{w} \in \mathbb{R}^{6}$ denotes the stacked linear and angular velocity of the EE frame origin w.r.t. the inertial frame, and $f_e$ is the desired contact force magnitude along $\bm{z}_E$ acting on the system from the environment, for details please refer to \cite{hui2023}.
\subsection{Static-Equilibrium Based Force Modeling}\label{sec:force model}
\begin{figure}[t]
      \centering
\includegraphics[width=\columnwidth]{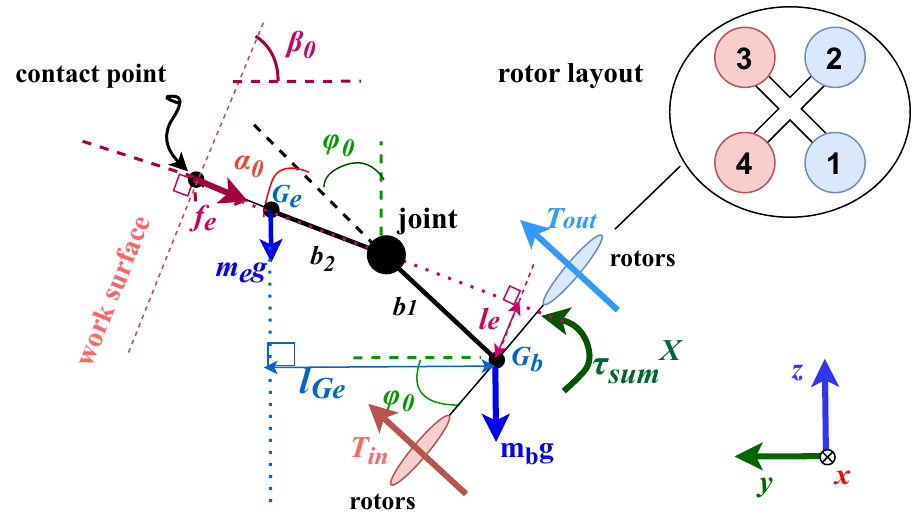}
      \caption{Free Body Diagram, $T_{in}$ is the thrust for the rotor pair close to the work surface, $T_{out}$ is the thrust for the other rotor pair further from the work surface, $\tau_{sum}^X$ is the total torque magnitude along $\bm{x}$ of the inertial frame.}
      \label{fig:fbd_new}
\end{figure}

As the previous work \cite{hui2023} claims, while achieving stable interactions, we assume that the whole system can be considered as a rigid body, resting at a quasi-static equilibrium phase. During the static equilibrium phase, zero net forces and torques are acting on the system CoM to maintain the equilibrium both in linear and angular dynamics in the 2-D plane, see the free body diagram in Fig.~\ref{fig:fbd_new}. We consider $\beta_0=|\beta^{w}|,\varphi_0=|\varphi^{w}|,\alpha_0=|\alpha^{w}|$ based on the symmetric property of the system. $\bm{G}_e$ and $\bm{G}_b$ are the CoM points of the manipulator and the aerial vehicle respectively. $m_b$ and $m_e$ are the mass of the aerial vehicle and the manipulator respectively. $l_e \in \mathbb{R}$ is the minimum distance between the desired contact force vector acting on the single contact point and the CoM of the aerial vehicle $\bm{G}_b$, and $l_{G_e} \in \mathbb{R}$ is the minimum distance between the gravity force vector of the manipulator acting on $\bm{G}_e$ and $\bm{G}_b$.

\begin{figure}[t]
      \centering     \includegraphics[width=\columnwidth]{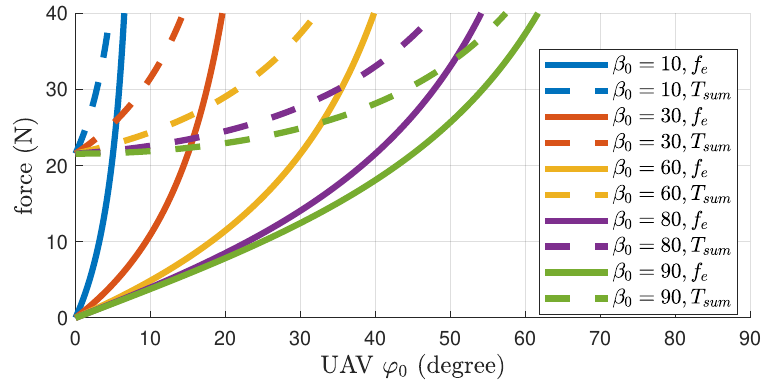}
      \caption{Predicted $f_e$ and $T_{sum}$ as in Eq.~(\ref{eq:fe})(\ref{eq:tsum}) associated with different $\varphi_0$ and $\beta_0$. They both increase along with increased $\varphi_0$ for a certain $\beta_0$.}
      \label{fig:fe}
\end{figure}

Considering the simplified system model in 2-D plane, when the system rests at the quasi static-equilibrium state, we have $|\bm{T}_1^{w}|=|\bm{T}_2^{w}|$ and $|\bm{T}_3^{w}|=|\bm{T}_4^{w}|$ assuming negligible uncertainties caused by modeling errors. Due to the symmetric property of the system, we define $T_{in} \in \mathbb{R}$ as the thrust magnitude of the rotor pair close to the work surface, and $T_{out} \in \mathbb{R}$ as the thrust magnitude of the rest pair of rotors. We define $T_{sum} \in \mathbb{R}$ as the total thrust magnitude of the 4 rotors, and $\tau_{sum}^X \in \mathbb{R}$ is the total torque magnitude of the 4 rotors along $\bm{x}$ axis. They are given by:
\begin{subequations}
 \begin{equation} \label{eq:ta}      T_{sum}=2(T_{in}+T_{out}),
 \end{equation}
 \begin{equation} \label{eq:taux}
     \tau_{sum}^X=2(T_{out}-T_{in})\cdot|\bm{r}^B|,
 \end{equation}
 \end{subequations}
 where $\bm{r}^B \in \mathbb{R}^3$ is the vector from the propeller center where the rotor is attached to the CoM of the aerial vehicle, neglecting the height difference between the propellers and the CoM of the aerial vehicle. With $f_e$ acting on the system along $\bm{z}_E$, assuming negligible friction force, the forces and torques acting on the system w.r.t. $\bm{G}_b$ can be written as:
 \begin{subequations}
\begin{equation}
    T_{sum}\cdot cos(\varphi_0)=m_eg+m_bg+f_e \cdot cos(\beta_0),
    \label{eq:z}
\end{equation}
\begin{equation}
    T_{sum} \cdot sin(\varphi_0)=f_e \cdot sin(\beta_0),
    \label{eq:y}
\end{equation}
\begin{equation}
    \tau_{sum}^X +m_eg \cdot l_{G_e}=f_e \cdot l_e.
    \label{eq:torque}
\end{equation}
\end{subequations}
where we define:
\begin{itemize}
    \item $\alpha_0 \neq 0$, $\beta_0 \neq 0$,
    \item $\varphi_0 < \beta_0$,
    \item $\beta^{w}, \varphi^{w}, \alpha^{w}$ always have the same sign such that the relation $\beta_0=\varphi_0+\alpha_0$ holds.
\end{itemize}
By re-arranging the Eq. (\ref{eq:z}) (\ref{eq:y}) with $\alpha_0=\beta_0-\varphi_0$, 
one can get:
\begin{align}
    &f_e=G_t\frac{sin(\varphi_0)}{sin(\beta_0-\varphi_0)},
    \label{eq:fe}\\
    &T_{sum}=G_t\frac{sin(\beta_0))}{sin(\beta_0-\varphi_0)},
    \label{eq:tsum}
\end{align}
where $G_t=m_eg+m_bg  \in \mathbb{R}$ is the total gravity force of the whole system. With Eq.~(\ref{eq:fe})(\ref{eq:tsum}), the predicted $f_e$ and $T_{sum}$ along with increased $\varphi_0$ for different surface orientation $\beta_0$ is shown in Fig.~\ref{fig:fe}. For a known work surface orientation $\beta_0$, the value of $f_e$ and $T_{sum}$ only corresponds to one unique value of $\varphi_0$ considering the task constraints in Eq.~(\ref{eq:task}). 

\section{Control and Hardware Design}
\label{sec:hardware_control}
In this section, we use an attitude controller for physical interactions based on the relation between the roll angle of the aerial vehicle and the interaction force value in the proposed force models in Eq.~(\ref{eq:fe}). Moreover, a manipulator design is introduced to ensure the successful integration of such control strategy even in the presence of uncertainties.
\subsection{Interaction Control}
The mathematical force model in Eq.~(\ref{eq:fe}) outlines the one-to-one correspondence between the aerial vehicle roll angle $\varphi_0$ and the desired interaction force magnitude $f_e$ considering a known work surface oriented by $\beta_0$. Therefore, the interaction force along the designated direction can be indirectly regulated by controlling the aerial vehicle attitude and the joint position. Considering a reference roll angle value of the aerial vehicle set to $\varphi_0$ and the joint position value set to $\alpha_0=\beta_0-\varphi_0$, the force value exerted from the platform is predictable during the interactions with the environment for the targeted task. During the free flight, a cascade position controller is applied~\cite{Lee2011} to reach a reference position close to the work surface. Once the platform reaches the reference position, the controller is switched to attitude control with the predefined attitude reference and joint position to approach and interact with the work surface. Consequently, highly accurate attitude control and state estimation are required to achieve the quasi-static equilibrium state described in Sec.\ref{sec:force model}. In practical cases, the accuracy of the controlled attitude can be affected by many factors, such as modeling errors, or environmental uncertainties. In order to address this challenge, we propose the following aerial manipulator design to ensure successful integration of the attitude control for the targeted interaction task.

\subsection{Design and Prototype of the Aerial Manipulator}\label{sec:design}
\begin{figure}[t]
      \centering
\begin{subfigure}[b]{0.6\columnwidth}
     \includegraphics[width=\columnwidth]{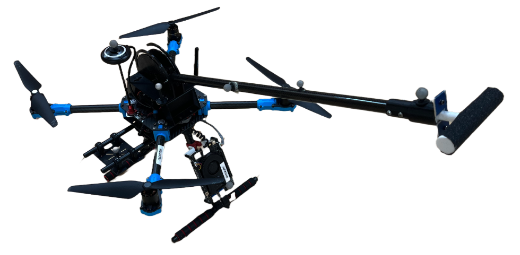}
      \caption{Aerial manipulator.}
      \label{fig:drone}
\end{subfigure} 
\hfill
\begin{subfigure}[b]{0.35\columnwidth}
     \includegraphics[width=\columnwidth]{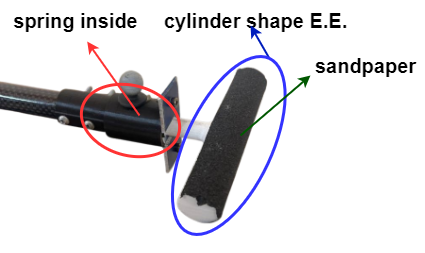}
      \caption{EE tip.}
      \label{fig:ee}
\end{subfigure} 
\begin{subfigure}[b]{0.5\columnwidth}
     \includegraphics[width=\columnwidth]{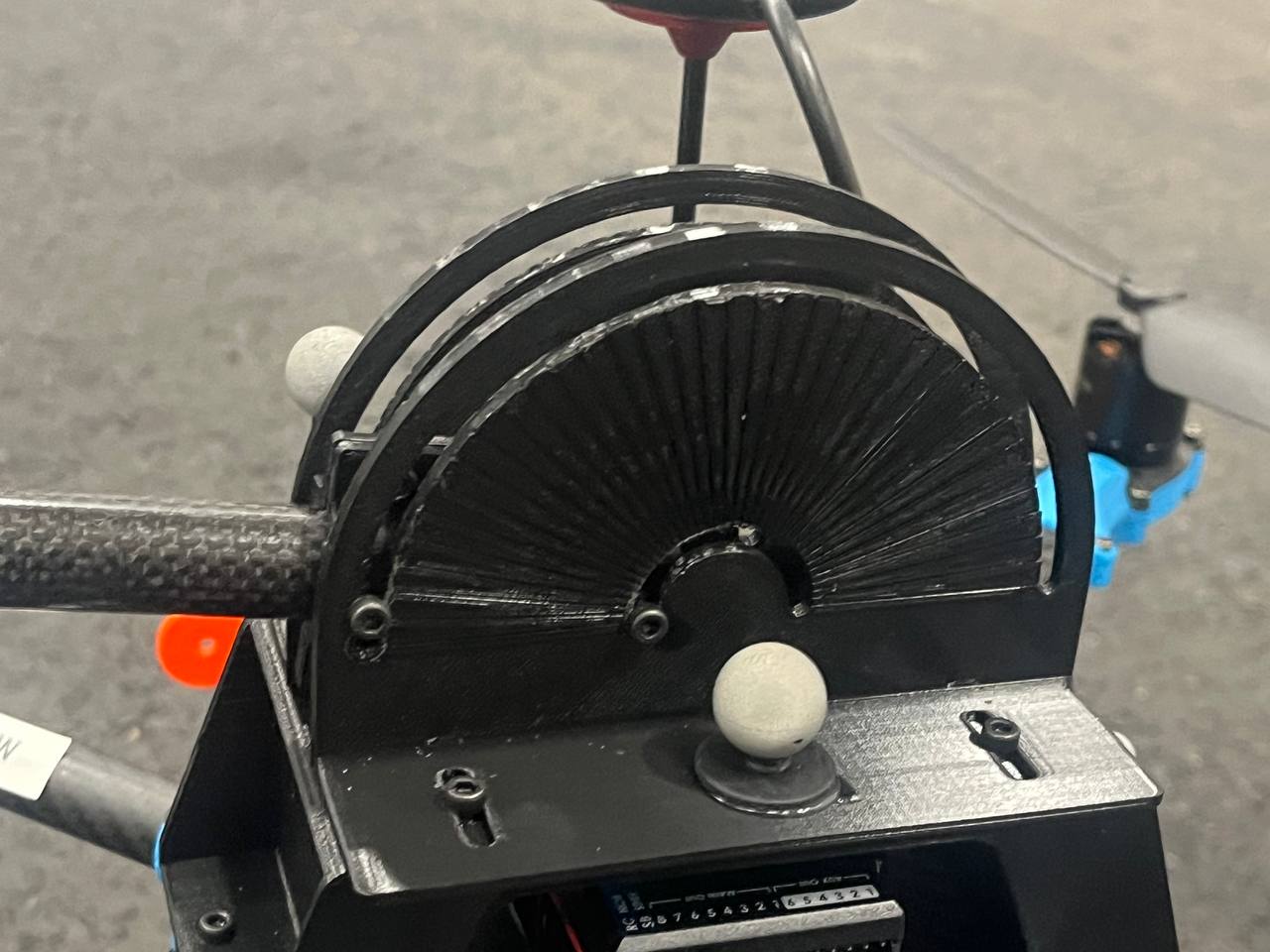}
      \caption{Revolute joint with locking screws.}
      \label{fig:joint}
\end{subfigure}
\hfill
\begin{subfigure}[b]{0.4\columnwidth}
     \includegraphics[width=\columnwidth]{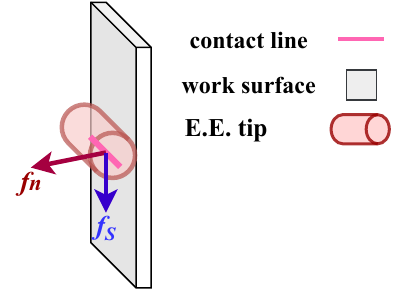}
      \caption{Cylindrical shape EE in contact with the work surface.}
      \label{fig:contact}
\end{subfigure}
\caption{The quadrotor-based aerial manipulator prototype developed for the targeted task.}
\label{fig:design}
\end{figure}

An aerial manipulator is developed according to the simplified system model in Sec.\ref{sec:bg}. An actuator positioned at the joint between the manipulator and the aerial vehicle is constrained in its torque generation capacity due to saturation. To overcome this limitation, a lockable revolute joint is employed to connect the manipulator to the aerial platform Holybro X500 V2, allowing 1-DoF rotation around the $\bm{x}_B$ axis of the body frame, as in Fig.~\ref{fig:drone}. It allows one to lock the revolute joint via screws at the preferred joint position for various testing conditions, see Fig.~\ref{fig:joint}. Once the joint is locked, the manipulator can be considered as rigidly attached to the aerial vehicle. Moreover, a spring is mounted below the EE tip to reduce the effects on the system caused by impact during the initial contact, see Fig.~\ref{fig:ee}. 

The EE tip has a cylindrical shape which forms a line contact with the flat work surface as in Fig.~\ref{fig:contact}. The interaction force vector from the surface towards the robot can be decomposed into two components $f_n, f_S \in \mathbb{R^+}$ acting at the CoP (Center of Pressure) point \cite{Sardain2004} which locates on the contact line. $f_n$ is normal to the surface while $f_S$ is the friction force tangential to the surface. The cylindrical shape design allows the platform to rotate around the contact line while the interaction force vector lies inside the friction cone, i.e. $f_S\leq \mu_S f_n$, where $\mu_S \in \mathbb{R}^+$ is the static friction coefficient. Without loss of generality, our EE design makes use of a sandpaper on the EE cylinder surface (see Fig.~\ref{fig:ee}) to ensure a high friction coefficient between the EE and a wooden work surface. However, other materials can also be employed. The friction condition ensures that the platform can rely on a static contact condition while regulating its roll angle during the transient to reach the task condition in Eq.~(\ref{eq:task}), which promises negligible friction force in the simplified 2-D plane. An EE tip with a spherical shape will allow rotations of the platform also in the directions outside the simplified 2-D plane which are undesired for the targeted task. Instead, the proposed cylindrical shape design has a long beam being perpendicular to the restricted 2-D plane and helps the E.E. tip fully align with the flat work surface with the formed contact line. The effectiveness of such design is validated through physical experiments.

\section{Safety Assessment}\label{sec:safe_eva}
In this section, we establish the safety assessment process applied to the developed aerial manipulator system for the targeted task. The force modeling on the thrust of each actuator is derived from Sec.~\ref{sec:force model} and used to predict the thrust level of each actuator w.r.t. its saturation. With the predicted thrust level of each actuator, we identify the safe and critical operation zones during physical interactions as introduced in Sec.\ref{sec:contribution}. Furthermore, the risk level of operations in the critical zone is evaluated to ensure safe operations when actuators work near saturation.
\subsection{Identification of Safe and Critical Interaction Zones}
In our previous work~\cite{hui2023}, singularity analysis is introduced based on the mathematical force models derived in Eq.~(\ref{eq:fe})(\ref{eq:tsum}) which outlines the critical cases of $T_{sum}$ considering the total actuation saturation. $T_{sum}$ however, only considers the total thrust condition of the actuators. With the definition of safe and operation zones in Sec.\ref{sec:contribution}, the thrust condition of each propeller of the aerial vehicle is more crucial for the safety assessment process. Therefore, from the quasi-static equilibrium state during stable interactions introduced in Sec.\ref{sec:bg}, the mathematical expressions of the thrust values of the 2 pairs of propellers $T_{in}$ and $T_{out}$ can be derived. By substituting Eq.~(\ref{eq:ta})(\ref{eq:taux}) and $\beta_0=\varphi_0+\alpha_0$ into Eq.~(\ref{eq:z})(\ref{eq:y})(\ref{eq:torque}), one has:
 \begin{align}
     &T_{out}=\frac{G_t \cdot sin(\beta_0)}{4 \cdot sin(\alpha_0)}+\Big(\frac{G_t l_e \cdot  sin(\varphi_0) }{2|r^B| \cdot sin(\alpha_0)}-\frac{m_eg \cdot l_{G_e}}{2|r^B|}\Big) \label{eq:tout},\\
     &T_{in}=\frac{G_t \cdot sin(\beta_0)}{4 \cdot sin(\alpha_0)}-\Big(\frac{G_t l_e \cdot  sin(\varphi_0) }{2|r^B| \cdot sin(\alpha_0)}-\frac{m_eg \cdot l_{G_e}}{2|r^B|}\Big) \label{eq:tin},
 \end{align}
where 
\begin{align*}
    &l_{G_e}=b_2 \cdot sin(\beta_0)+b_1 \cdot sin(\varphi_0),\\
    &l_e=b_1 \cdot sin(\alpha_0).
\end{align*}
The parameters related to the aerial manipulator described in Sec.~\ref{sec:design} are $m_b=2.1$\si{\kilo\gram}, $m_e=0.1$\si{\kilo\gram}, $|\bm{r}^B|=0.266$\si{\meter}, $b_1=0.113$\si{\meter}, $b_2=0.593$\si{\meter}. The maximum thrust of the aerial vehicle was estimated testing the hovering state along free flights. During the hovering test, the total thrust ratio is $C_h=\displaystyle\frac{T_{hover}}{T_{sum}^{max}}=0.61$ with $T_{hover}=(m_b+m_e)g=G_t$, and $g=9.8$\si{\meter\per\square\second}. The estimated maximum total thrust and the single thrust of each propeller are:
\begin{equation*}
   T_{sum}^{max}=\frac{G_t}{C_h}=35.2\si{\newton}, \ T_{max}=\frac{T_{sum}^{max}}{ 4}=8.8\si{\newton}.
\end{equation*}
With the identified parameters and $T_{max}$, Eq.~(\ref{eq:tout})(\ref{eq:tin}) can be displayed for $\beta_0=|\beta^{w}| \in (0\degree,90\degree]$ as in Fig.~\ref{fig:zones}. 

Under a certain value of $\beta_0$, we identify the safe interaction zone as the feasible range of $\varphi_0$ of the aerial vehicle within which the desired thrust of the interaction task does not exceed the saturation multiplying a safety factor $\eta \in (0,0.9]$. The safety factor takes into account uncertainties during the physical interaction. And the critical interaction zone is identified as the range of $\varphi_0$, where the desired thrust is higher than the saturation multiplying $\eta$ but still within the saturation. In the critical zone, the platform has higher chances to reach saturation when subjected to uncertainties since the actuators are working near saturation. The area above the critical zone exceeds the platform saturation, and operations in this area will result into failure of the interaction task and even cause instability of the platform. In Fig.~\ref{fig:zones}, the two dashed lines $T_{max}$ and $\eta T_{max}$ where $\eta=0.7$ are displayed as the boundary lines to assess the safe and critical interaction zones. $T_{out}$ and $T_{in}$ have almost the same values when $\beta_0$ is small. However, $T_{out}$ reaches the boundary lines first along with increased $\varphi_0$ under bigger value of $\beta_0$ which is considered more crucial w.r.t. $T_{in}$. Consequently, $T_{out}$ is used to identify the range of $\varphi_0$ for the two interaction zones. The identified safe and critical interaction zones for diverse oriented working surfaces are shown in Table~\ref{table:1}. For a specific surface orientation $\beta_0$, each zone is formed by an upper boundary and a lower boundary of the roll angle magnitude $\varphi_0$ of the aerial vehicle.
\begin{table}\caption{safe and critical interaction zones}\label{table:1}
\begin{center}      
    \begin{tabular}{|m{1cm}|m{2cm}|m{3cm}|} 
     \hline
     $\beta_0 (\degree)$ &Safe Zone $\varphi_0 (\degree)$ &Critical Zone $\varphi_0 (\degree)$ \\ 
     \hline
     $10$  &\cellcolor[HTML]{AAACED}$ 0\rightarrow1$  & \cellcolor[HTML]{AAACED}$2\rightarrow4$\\
     \hline
     $30$  &\cellcolor[HTML]{AAACED}$ 0\rightarrow3$  &\cellcolor[HTML]{AAACED} $4\rightarrow10$\\
     \hline
     $60$  &\cellcolor[HTML]{AAACED}$ 0\rightarrow7$ & \cellcolor[HTML]{AAACED} $8\rightarrow21$\\
     \hline
     $80$  &\cellcolor[HTML]{AAACED}$ 0\rightarrow10$ &\cellcolor[HTML]{AAACED} $11\rightarrow29$\\
     \hline
     $90$  &\cellcolor[HTML]{AAACED}$ 0\rightarrow12$ &\cellcolor[HTML]{AAACED} $13\rightarrow34$\\
     \hline
\end{tabular}
\end{center}
\end{table}
\subsection{Risk Assessment in the Critical Zone}
\begin{figure}[t]
      \centering
\includegraphics[width=\columnwidth]{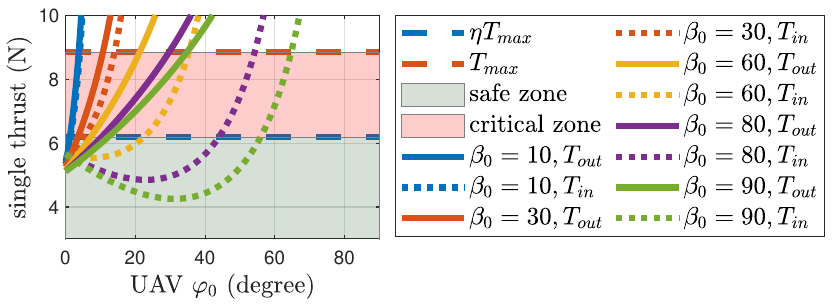}
      \caption{safe and critical interaction zones assessed by $\eta T_{max}$ and $T_{max}$, $\eta=0.7$ is the safety factor, $T_{max}$ is the thrust saturation of each propeller.}
      \label{fig:zones}
\end{figure}

In the critical zone, the thrust magnitude increases with a near linear slope w.r.t. the value of $\varphi_0$, as in Fig.~\ref{fig:zones}. Since the platform is operating with its actuators working near saturation, the system can be easily affected by uncertainties from modeling inaccuracy and the environment. In order to identify the feasible operation range of the platform against uncertainties, the risk level of different operation conditions in the critical zone is assessed. Assuming linear relationship between $\varphi_0$ and thrust in this zone, we propose a risk indicator $\lambda$ to evaluate the risk level of the operation cases in the critical zone. We define $\varphi_l, \varphi_u$ as the lower and upper boundaries of the roll angle in the critical zone for a specific work surface orientation $\beta_0$. Knowing $\varphi_l \rightarrow \varphi_u$, pushing on a surface oriented by $\beta_0$ with a roll angle $\varphi_0\geq \varphi_l$ has a risk level of:
\begin{equation}
    \lambda=\frac{\varphi_0}{\varphi_u} \leq 1,
\end{equation}
with higher $\lambda$ representing higher risk. 

The identified safe and critical interaction zones and the risk indicator for critical operations are used as guidelines for planning and executing experiments in the following section, with an objective of safely validating the force model in Eq.~(\ref{eq:fe}).

\section{Experiments}
\label{sec:exp}
In this section, we planned a series of experiments with different $(\beta_0, \varphi_0)$ which are evaluated by following the safety assessment results from the last section as in Fig.~\ref{fig:evaluation}. We aimed at validating the static-equilibrium based force model in Eq.~(\ref{eq:fe}) with safe physical interaction operations using the developed quadrotor-based aerial manipulator. 
\begin{figure}[!t]
      \centering
\begin{subfigure}[b]{0.4\columnwidth}
    \includegraphics[width=\columnwidth]{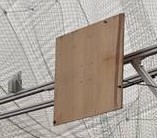}
      \caption{}
      \label{fig:setup}
\end{subfigure} 
\hfill
\begin{subfigure}[b]{0.4\columnwidth}
     \includegraphics[width=\columnwidth]{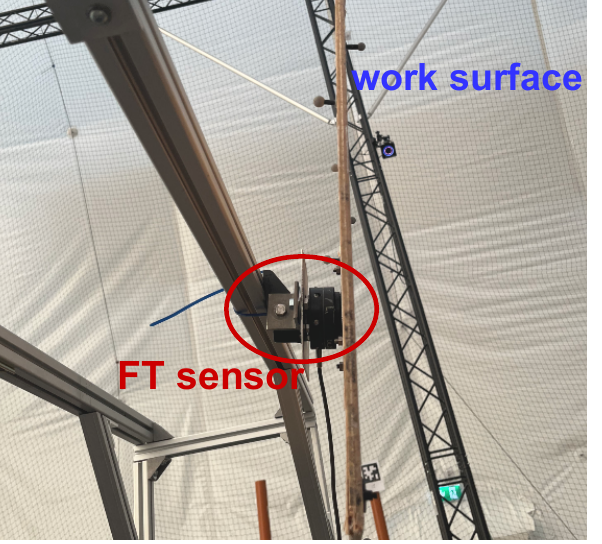}
      \caption{}
      \label{fig:force_sensor}
\end{subfigure}
\caption{Experiment setup, (a) a wooden board as work surface, (b) a FT sensor mounted behind the work surface.}
\label{fig:full_setup}
\end{figure}
\subsection{Experiment Setup}
\begin{figure}[!t]
      \centering
      \begin{subfigure}[b]{0.8\columnwidth}
\includegraphics[width=\columnwidth]{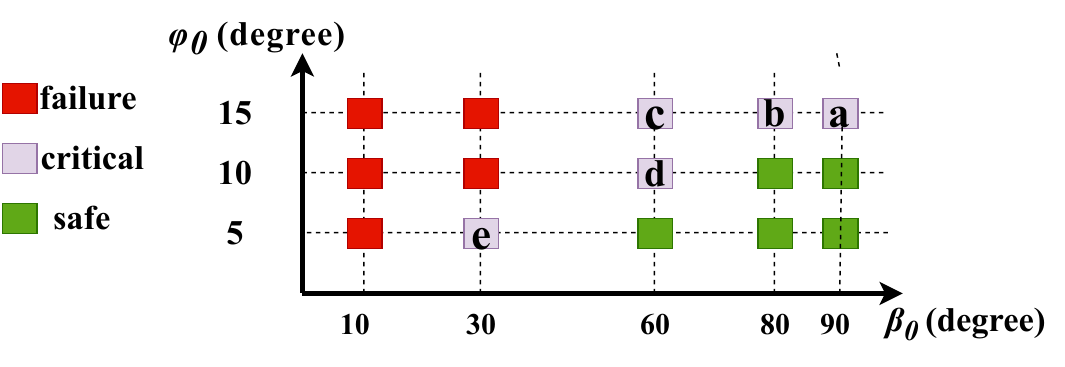}
      \caption{}
      \label{fig:cases}
      \end{subfigure}
      \hfill
   \begin{subfigure}[b]{0.15\columnwidth}
\includegraphics[width=\columnwidth]{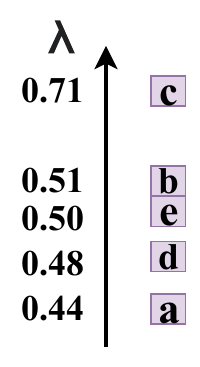}
      \caption{}
      \label{fig:risk}
      \end{subfigure} 
      \caption{Safety assessment of planned experiments using the interaction zones defined in Table~\ref{table:1}. (a): Operation cases identified as safe or critical cases. (b): Risk level of critical cases.}
      \label{fig:evaluation}
\end{figure}
\begin{figure}[!t]
      \centering
      \begin{subfigure}[b]{\columnwidth}\label{fig:U5}
\includegraphics[width=\columnwidth]{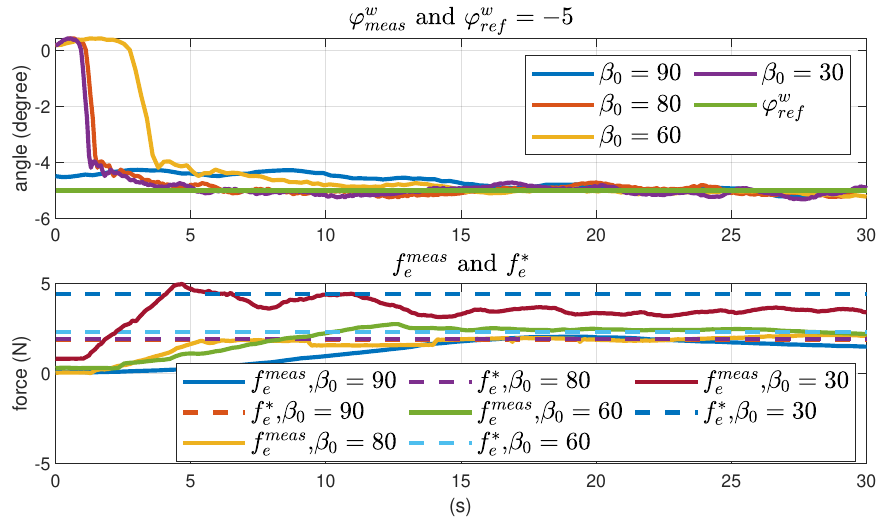}
 \caption{ $\varphi_{ref}^{w}=-5\degree$, $\beta_0=90\degree, 80\degree, 60\degree, 30\degree$.}
      \end{subfigure}
   \begin{subfigure}[b]{\columnwidth}
\includegraphics[width=\columnwidth]{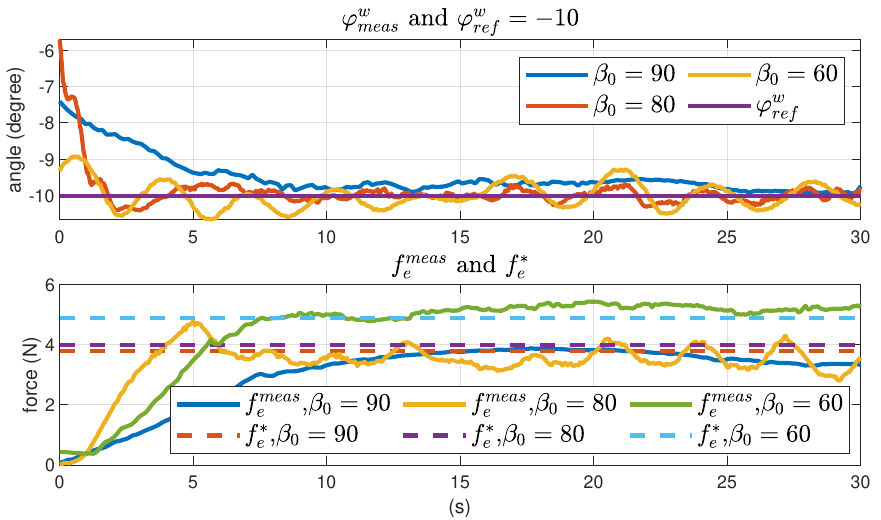}
      \caption{ $\varphi_{ref}^{w}=-10\degree$, $\beta_0=90\degree, 80\degree, 60\degree$.}
      \label{fig:U10}
      \end{subfigure} 
 \begin{subfigure}[b]{\columnwidth}
\includegraphics[width=\columnwidth]{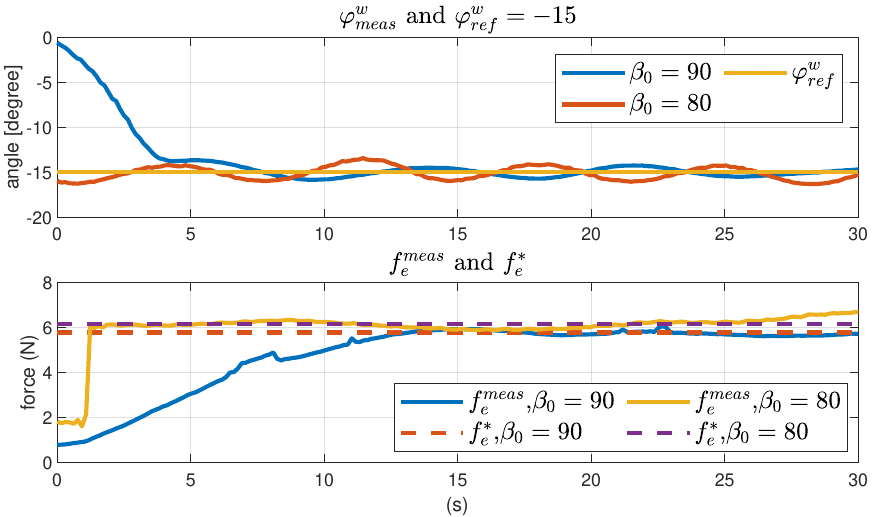}
\caption{$\varphi_{ref}^{w}=-15\degree$, $\beta_0=90\degree, 80\degree$.}
      \label{fig:U15}
      \end{subfigure} 
      \caption{Measured roll angle $\varphi_{meas}^{w}$ of the aerial vehicle, measured interaction force component perpendicular to the work surface $f_e^{meas}$, reference roll angle $\varphi_{ref}^{w}$, predicted force value $f_e^*$. }
      \label{fig:plots}
\end{figure}
The aerial manipulator developed in Sec.\ref{sec:hardware_control} includes a Pixhawk autopilot and a computer onboard to manage the physical interaction operation. An Optitrack as Motion Capture (MoCap) system is used to receive position and orientation data during the experiments. A wooden board is fixed on a crane as the work surface with adjustable orientation to obtain different value of $\beta_0$ as in Fig.~\ref{fig:setup}. Moreover, a 6-DoF FT (force and torque) sensor is mounted between the board and the crane to obtain the ground truth at contact during interactions as in Fig.~\ref{fig:force_sensor} for validating the force model in Eq.~(\ref{eq:fe}). The MoCap system is also used to set the wooden board angle $\beta_0$. 

To validate the force models, $\beta_0=10\degree, 30\degree, 60\degree, 80\degree, 90\degree$ are planned for testing with $\varphi_0=5 \degree, 10\degree, 15\degree$ (during the experiments both $\beta^w$ and $\varphi^w$ are negative). Based on the safe and critical interaction zones identified in Table~\ref{table:1} from the previous section, operation cases with different $(\beta_0,\varphi_0)$ are evaluated as in Fig.~\ref{fig:cases}. Moreover, the risk level $\lambda$ of the critical cases are evaluated as in Fig.~\ref{fig:risk}. The experiments are planned regarding to the safety assessment results. All the experiments identified as safe operations were executed before the critical operations. The execution of critical cases followed the order from the bottom to the top of Fig.~\ref{fig:risk}. The execution was stopped once task failure occurred to avoid potential platform damage in operations with even higher risk level. The identified failure cases are not executed to prevent unnecessary crashes or risky situations.
\subsection{Experiment Results}
The operation cases identified as safe cases in Fig.~\ref{fig:cases} were all executed successfully for the targeted interaction task. Among all the planned operation cases in the critical zone in Fig.~\ref{fig:risk}, the platform was not able to reach the desired quasi static-equilibrium state under the case when $\beta_0=60\degree$, $\varphi_0=15\degree$ which has the highest risk level of $\lambda=0.71$. The data of the successfully executed operations during the experiments are presented in Fig.~\ref{fig:plots}.
Measured roll angle of the aerial vehicle expressed in the inertial frame $\varphi_{meas}^{w}$ and measured force component perpendicular to the work surface $f_e^{meas}$ are displayed for the reference roll angle of $\varphi_{ref}^{w}=-5 \degree$, $-10 \degree$, $-15\degree$ respectively. The predicted force values $f_e^*$ from the static-equilibrium based modeling in Eq.~(\ref{eq:fe}) are shown as comparison for validating the proposed force model with acceptable errors between the $f_e^{meas}$ and $f_e^*$. With the above listed experiments, we safely demonstrated the static-equilibrium based force modeling approach via experiments following the guidelines from the safety assessment process. The experiment of the failed critical case is shown in the video attachment. From the experiment results, for our particular platform and setup, we suggest to keep the physical interaction operations within the identified safe interaction zone. If operations in the critical zone are required, we suggest to keep the risk level $\lambda$ within $0.5$ for safe and successful physical interactions and execute with protections to avoid platform damage. The video featuring the executed experiments is available at \url{https://youtu.be/Nn_ZJ1rVCSE}.

\section{Conclusion}
\label{sec:con}
In this paper, we presented a framework for applying a safety assessment process to ensure safe pushing on diverse oriented work surfaces using an underactuated aerial vehicle. The process involved evaluating the predicted thrust level of each actuator w.r.t. its saturation during pushing. With the safety assessment results, we validated a static-equilibrium based force modeling approach via practical experiments across operations with variable surface orientation. This work provides guidelines in safe utilization of underactuated aerial vehicles for contact-based inspections on variably oriented work surfaces in industrial applications.


\addtolength{\textheight}{-12cm}   






% ...

\end{document}